\documentclass{article}

\PassOptionsToPackage{numbers, compress}{natbib}

\usepackage[preprint]{neurips_2025}

\usepackage{xspace}
\usepackage{amsmath} 

\makeatletter
\DeclareRobustCommand\onedot{\futurelet\@let@token\@onedot}
\def\@onedot{\ifx\@let@token.\else.\null\fi\xspace}
\def\eg{\emph{e.g}\onedot}

\makeatother

\usepackage[utf8]{inputenc} 
\usepackage[T1]{fontenc}    
\usepackage{xcolor}  
\definecolor{iccvblue}{rgb}{0.21,0.49,0.74}
\usepackage[pagebackref,breaklinks,colorlinks,allcolors=iccvblue]{hyperref}
\usepackage{url}            
\usepackage{booktabs}       
\usepackage{amsfonts}       
\usepackage{nicefrac}       
\usepackage{microtype}      

\usepackage{array}
\usepackage{lipsum} 
\usepackage{graphicx}
\usepackage{enumitem}
\usepackage{multirow}
\usepackage{multicol}
\usepackage{subcaption}
\usepackage{wrapfig}

\usepackage{enumitem}
\usepackage{mathtools}
\usepackage{ragged2e}
\usepackage{xpatch}
\usepackage{graphicx}
\usepackage{soul}
\usepackage[normalem]{ulem}
\usepackage{pifont}
\usepackage{algorithm}
\usepackage{algorithmic}
\usepackage{amsmath}

\usepackage{pifont}
\newcommand{\cmark}{\ding{51}}%
\newcommand{\xmark}{\ding{55}}%

\definecolor{lightgray2}{rgb}{0.88,0.88,0.88}
\newcommand{\trf}[1]{{\textbf{\color{red}{#1}}}} %
\newcommand{\tbd}[1]{{\color{blue}{\underline{#1}}}} %

\title{RAGSR: Regional Attention Guided Diffusion for Image Super-Resolution}

\author{
Haodong He$^{1,2*}$, 
Yancheng Bai$^{2*}$\footnotemark[3] \space,  
Rui Lan$^{2}$, Xu Duan$^{2}$, \\
\textbf{Lei Sun}$^{2}$\footnotemark[2] \space \textbf{, Xiangxiang Chu}$^{2}$\textbf{, and Gui-Song Xia}$^{1}$\footnotemark[2] \\
{$^{1}$School of Computer Science, Wuhan University \qquad $^{2}$Amap, Alibaba Group}
}

\begin{document}

\maketitle

\begin{figure}[h]
    \centering
    \includegraphics[width=\linewidth]{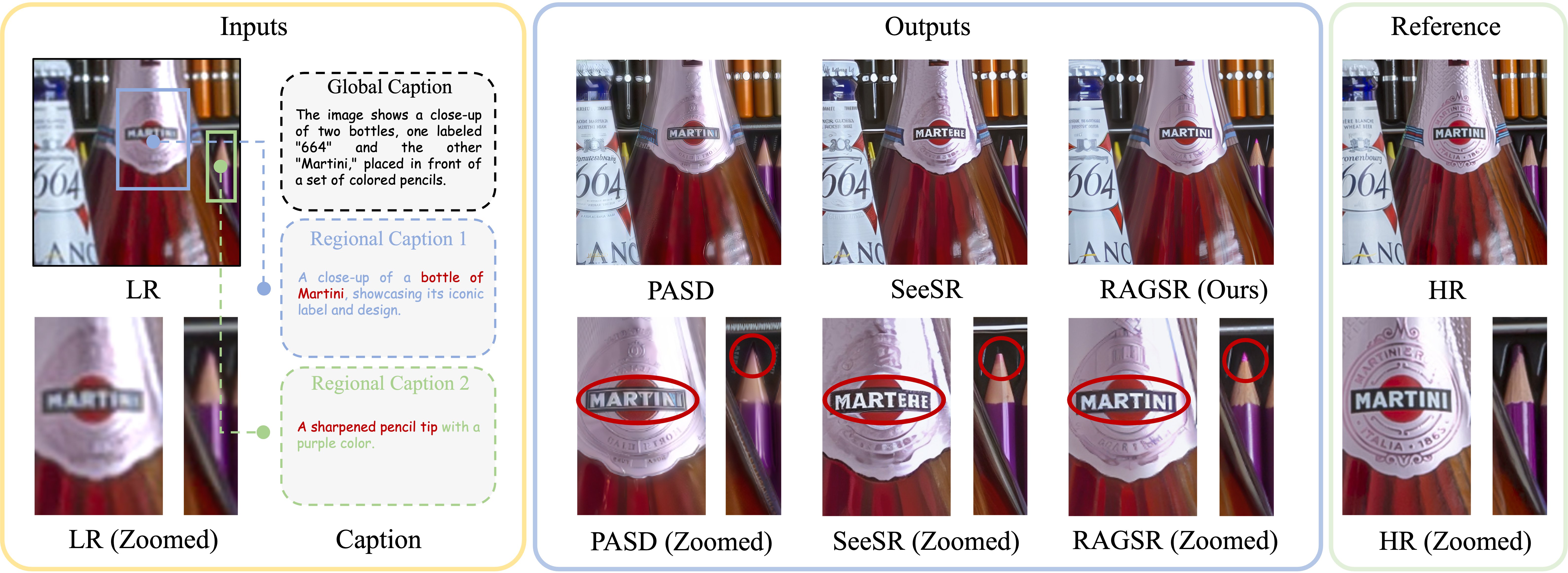}
    \caption{RAGSR leverages both global textual captions and regional textual captions to infuse low-resolution (LR) images with rich semantic information. By enforcing explicit alignment between textual descriptions and image regions, RAGSR enables more precise guidance for reconstructing localized image details, thereby surpassing state-of-the-art approaches.}
    \label{fig:motivation}
\end{figure}

\renewcommand{\thefootnote}{\fnsymbol{footnote}}

\footnotetext[1]{Equal Contribution.}
\footnotetext[2]{Corresponding Authors.}
\footnotetext[3]{Project Leader.}

\begin{abstract}
The rich textual information of large vision-language models (VLMs) combined with the powerful generative prior of pre-trained text-to-image (T2I) diffusion models has achieved impressive performance in single-image super-resolution (SISR). 
However, existing methods still face significant challenges in generating clear and accurate regional details, particularly in scenarios involving multiple objects. This challenge primarily stems from a lack of fine-grained regional descriptions and the models' insufficient ability to capture complex prompts. 
To address these limitations, we propose a Regional Attention Guided Super-Resolution (RAGSR) method that explicitly extracts localized fine-grained information and effectively encodes it through a novel regional attention mechanism, enabling both enhanced detail and overall visually coherent SR results.
Specifically, RAGSR localizes object regions in an image and assigns fine-grained caption to each region, which are formatted as region-text pairs as textual priors for T2I models. A regional guided attention is then leveraged to ensure that each region-text pair is properly considered in the attention process while preventing unwanted interactions between unrelated region-text pairs. By leveraging this attention mechanism, our approach offers finer control over the integration of text and image information, thereby effectively overcoming limitations faced by traditional SISR techniques. Experimental results on benchmark datasets demonstrate that our approach exhibits superior performance in generating perceptually authentic visual details while maintaining contextual consistency compared to existing approaches.
\end{abstract}

\section{Introduction}
The core objective of single image super-resolution (SISR) is to reconstruct a high-resolution (HR) image from a low-resolution (LR) input while preserving semantic identity and texture details. Real-world image degradation (\eg, compression, blur, and noise) poses significant challenges for this task. Traditional methods typically treat SISR as a regression problem and use a deep neural network to learn a direct mapping from LR images to HR ones \cite{dong2015image, wang2018esrgan, li2018multi}. These regression-based methods often achieve high scores on paired metrics such as PSNR and SSIM \cite{ssim}, however, the generated results are over-smooth and of degraded quality comparable to natural images. Adversarial training enables GAN-based techniques like BSRGAN \cite{BSRGAN2021} and Real-ESRGAN \cite{RealESRGAN2021} to produce realistic textures by capturing stochastic degradation processes. In contrast, they suffer from training instability, and their outputs often exhibit unnatural visual artifacts. Subsequent works, such as LDL \cite{LDL2022} and DeSRA \cite{DeSRA2023}, mitigate many of these artifacts but struggle to generate more natural details. Recently, the powerful generative models, such as diffusion models \cite{rombach2022high,podell2023sdxl,song2023fashion,zeng2024cat,lan2025fluxtext} and auto-regressive models \cite{gpt_chen20s,xu2025scalar}, have revolutionized image generation tasks, including text-to-image (T2I) synthesis, which has inspired recent efforts to leverage the pre-trained T2I models for downstream tasks like SISR \cite{chen2020generative}. For instance, seminal works such as PASD \cite{PASD2024} leverage concise textual descriptions as prompts to guide T2I models in generating HR images. In contrast, SeeSR \cite{seesr2024} adopts semantic tags rather than captions as the input for T2I models. Recent studies \cite{qu2024xpsr,yu2024scaling} have further explored the use of large vision-language models (VLMs) to generate global textual descriptions from LR images, which are then utilized as prompts to synthesize HR images. 

Injecting rich textual priors into T2I models can significantly enhance the quality of generated high-resolution images. However, current methods still face challenges in improving localized details, particularly regarding fine-grained objects. We identify two primary challenges:
(1) \textbf{Limited capability in capturing complex textual descriptions}: Current T2I models struggle to effectively interpret and generate images from rich text prompts, especially when multiple objects are involved. As illustrated in Figure \ref{fig:motivation}, the PASD model \cite{PASD2024} successfully generates satisfactory results for a bottle when guided by a detailed global textual prompt. However, it produces a blurry image of a pencil due to misalignment between the text and the generated image. In contrast, the MMSR model \cite{mmsr2025} attempts to address this issue by integrating global text prompts with multi-modal visual cues such as semantic segmentation, depth maps, and edge maps. While this approach can improve performance, it incurs significant computational overhead and still struggles to deliver the desired detail in generation.
(2) \textbf{Lack of regional fine-grained descriptions}: Fine-grained descriptions are critical information for enhancing detail reconstruction. However, SeeSR \cite{seesr2024} uses semantic tags instead of descriptive captions. As shown in Figure \ref{fig:motivation}, SeeSR \cite{seesr2024} generates a globally plausible HR image using the tag ``wine bottle'', yet the super-resolved label on the bottle incorrectly appears as ``MARTEHE'', highlighting the limitations of tag-based conditioning.

To address these challenges, we propose a \textbf{R}egional \textbf{A}ttention \textbf{G}uided \textbf{SR} (\textbf{RAGSR}), which extracts explicit regional fine-grained captions through a VLM and a detection model and effectively injects this localized rich information into the model through a novel regional attention mechanism.  
Specifically, RAGSR consists of two stages. In Stage 1, we fine-tune a large vision-language model (\eg, Qwen2.5-VL \cite{qwen252025}) on LR images to extract accurate captions from degraded LR inputs. Subsequently, the open-vocabulary detection model (\eg, LLMDet \cite{fu2025llmdet}) is employed to identify foreground regions in LR images, and the fine-tuned Qwen2.5-VL generates corresponding captions for each detected region. In Stage 2, these region-caption pairs are integrated as semantic conditions into the T2I model to guide the super-resolution process. To ensure that each caption interacts exclusively with its associated region while suppressing interference from unrelated region-caption pairs, we introduce a regional attention mechanism that explicitly aligns textual descriptions with spatial regions through regional attention masks. The derived regional features will interact with global features, thereby enhancing the model’s ability to reconstruct fine-grained details while preserving the overall image quality.

Experimental results demonstrate that RAGSR achieves state-of-the-art performance on benchmark datasets such as DIV2K-val \cite{div2k}, RealSR \cite{realsr}, and DrealSR \cite{drealsr}. Quantitative evaluations further confirm its superiority over baseline methods, with significant improvements in both pixel-level fidelity and perceptual naturalness. This dual optimization highlights RAGSR’s effectiveness in balancing structural accuracy and semantic coherence, ensuring high-quality super-resolution outputs with reduced artifacts and enhanced detail preservation.

\section{Related Work}

\paragraph{Generative Models for SISR.} Pioneering works like SRCNN \cite{SRCNN2014} established deep learning for SISR, yet early methods \cite{secondorder2019,residualdensenetwork2018,imagesuper-resolution2018,enhanceddeepresidualnetwork2017,chu2020multi,efficientlone-rangeattention2022,chu2021fast,dualaggregationtransformerimage2023,activatingpixelsimagesuperresolution2023} constrained by idealized bicubic degradation assumptions, leading to limited generalization capability in real-world complex degradation scenarios. After that, generative models have become popular to realistic SISR, encompassing GAN-based \cite{goodfellow2020generative} approaches and diffusion model-based methods, each proposing distinct solutions for perceptual quality and detail preservation. GAN-based methods, such as BSRGAN \cite{BSRGAN2021} and Real-ESRGAN \cite{RealESRGAN2021}, generate high-fidelity textures by adversarial training to model stochastic degradation processes. However, these methods often exhibit training instability, leading to outputs with unnatural artifacts. Subsequent works, including LDL \cite{LDL2022} and DeSRA \cite{DeSRA2023}, attempt to mitigate these artifacts but at the cost of diminished detail generation capacity. Recently, researchers have adopted powerful pre-trained T2I models, such as Stable Diffusion \cite{LDM2022}, to address real-world ISR challenges. These models, trained on billions of image-text pairs, leverage robust image priors to tackle SISR tasks. For example, StableSR \cite{StableSR2024} achieves this goal by fine-tuning the SD model with a temporal-aware encoder and employing feature deformation to balance fidelity and perceptual quality. DiffBIR \cite{diffbir2024} adopts a two-stage strategy for SISR: it first reconstructs an image as an initial estimate and then enhances image details using SD prior to refinement.

\paragraph{Injecting Additional Information for SISR.} Recent works have focused on leveraging powerful T2I diffusion models by utilizing text prompts to fully exploit learned image priors \cite{qu2024xpsr,yu2024scaling}. PASD \cite{PASD2024} attempts to combine multi-modal guidance via YOLO \cite{redmon2016you} detectors and BLIP captions, but is constrained by the limited adaptability of detection models to degraded LR inputs and incomplete scene descriptions generated by BLIP. SeeSR \cite{seesr2024} introduces a fine-tuned RAM \cite{zhang2024recognize} to generate global captions describing LR images, thereby improving the quality of generated images. However, its tag-style captions lack rich semantic information critical for guiding the super-resolution process. In contrast, our method employs a Qwen2.5-VL model fine-tuned on LR images to generate more detailed captions, further unlocking the generative capacity of T2I models. 
Beyond caption quality, spatial-semantic ambiguity in text prompts poses challenges for SR tasks. To address this, MMSR \cite{mmsr2025} incorporates depth maps, segmentation maps, and edge maps to implicitly align spatial and textual information. However, this approach overlooks two key issues: (1) the performance degradation of modality extractors (\eg, Mask2Former \cite{mask2former2022}) when applied to LR inputs; and (2) the adverse effects of implicit alignment when modalities are missing. Our method explicitly aligns captions with their corresponding image regions, enhancing the model's ability to reconstruct localized details.

\section{Method}

\begin{figure}[t]
    \centering
    \includegraphics[width=\linewidth]{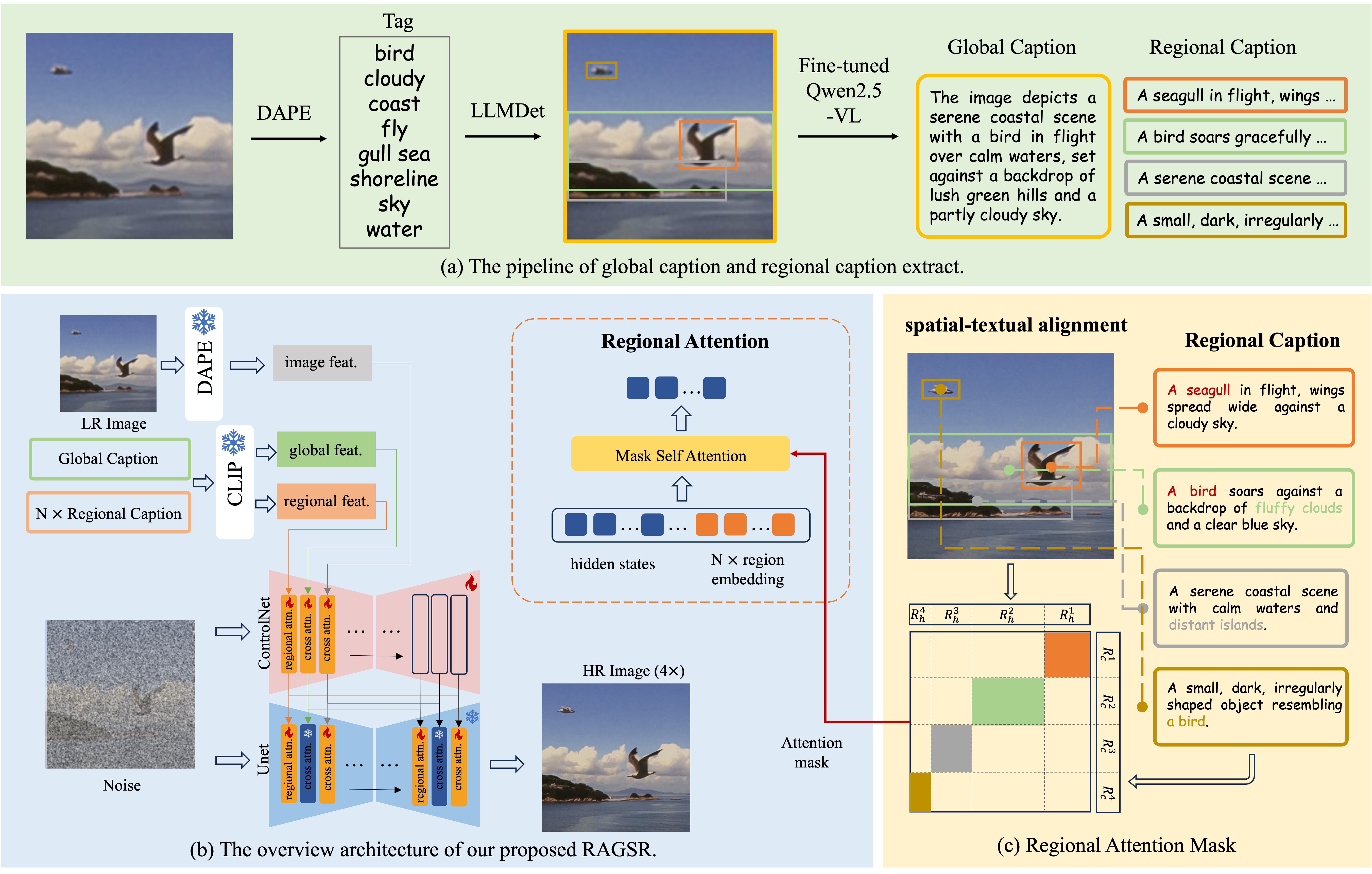}
    \caption{Overview of RAGSR. (a) The framework for fine-grained information extraction. We utilize DAPE~\cite{seesr2024} to obtain tags for low-resolution images, followed by the application of LLMDet to generate detection boxes with open vocabulary. Finally, we employ the fine-tuned Qwen2.5-VL model to obtain both global and regional captions. (b) The network architecture of RAGSR. The embeddings extracted from DAPE serve as conditions in the ControlNet. The global and regional embeddings obtained from CLIP provide fine-grained prompt information, which is input into the Regional Attention module to guide the details of the generated images. (c) Illustration of the regional attention mask. In the self-attention module, the specified region's information interacts with the corresponding regional caption.}
    \label{fig:ragsr}
\end{figure}

Reconstructing HR images from information-scarce LR inputs is an ill-posed task, requiring external priors or assumptions to bridge the information gap. Inspired by prior works, we propose to extract high-quality global and regional descriptions from LR images and explicitly align them with corresponding image regions to achieve superior super-resolution performance. Our proposed model, RAGSR, is illustrated in Figure \ref{fig:ragsr}. The training process comprises two stages: in the first stage, we train a robust Qwen2.5-VL \cite{qwen252025} to extract high-quality descriptions from degraded LR images. Specifically, we input degraded LR images and optimize the model to minimize the discrepancy between generated text and ground truth annotations during training. As shown in Figure \ref{fig:ragsr} (a), the fine-tuned Qwen2.5-VL takes the detected regions from LLMDet \cite{fu2025llmdet} on LR images to generate high-quality global and regional descriptions. These detected bounding boxes and their corresponding high-quality descriptions serve as conditional priors into the T2I model. In the second stage, as shown in Figure \ref{fig:ragsr} (b), we employ a regional attention mechanism to align the textual descriptions with their corresponding image regions while integrating global semantic information, enabling the generation of visually pleasing and semantically accurate SISR results.

\subsection{Region-aware Caption Generation}

Most VLMs are trained and evaluated on clean datasets, often suffering severe performance degradation when applied to visually impaired LR images. This limitation results in erroneous semantic information being extracted from LR inputs, which can mislead T2I models into generating semantically inconsistent super-resolution results \cite{seesr2024}. To optimize the model's performance in generating region-aware descriptions under degraded conditions, we fine-tuned the Qwen2.5-VL \cite{qwen252025} architecture using synthetically generated LR images to enhance its robustness against visual degradation. 

\begin{figure}[h]
    \centering
    \includegraphics[width=\linewidth]{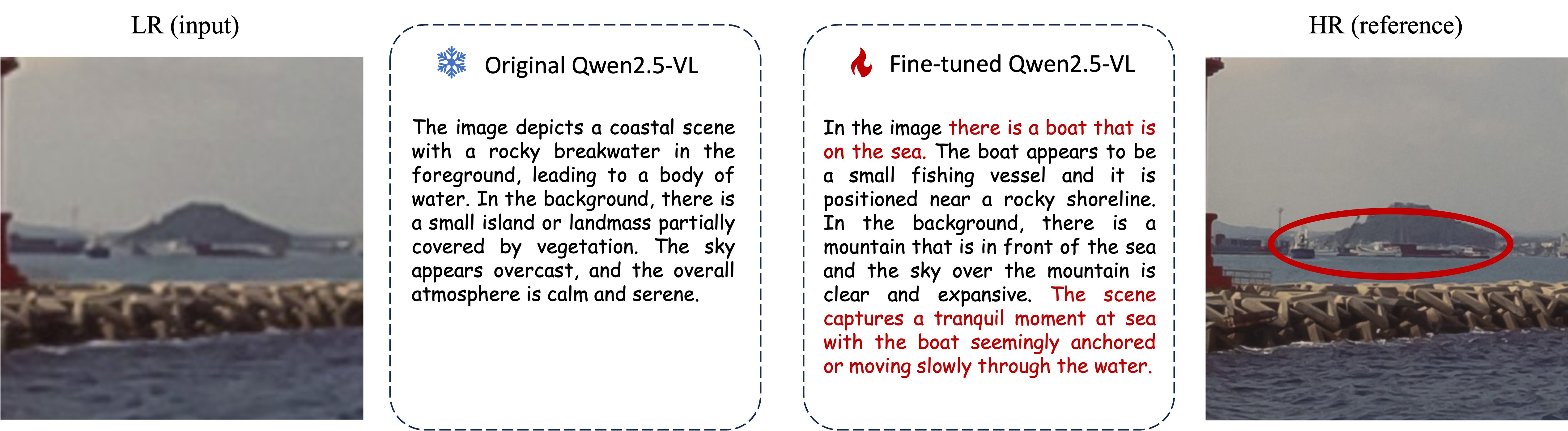}
    \caption{The outputs of Qwen2.5-VL with and without fine-tuning. The fine-tuned model extracts more precise information from degraded LR images, such as identifying ``boat''. }
    \label{fig:caption}
\end{figure}

Specifically, we first apply random degradation to HR images to obtain their corresponding LR counterparts $x$. These degraded images $x$ and their captions are then used to fine-tune the Qwen2.5-VL model, which is optimized to minimize the discrepancy between the generated captions and ground-truth annotations. The training objective is formulated as:

\begin{equation}
    L_{\text{Qwen}} = {CE}(f(x), GT)
\end{equation}

where ${CE}$ denotes the cross-entropy loss and $f(x)$ means the model's outputs on LR images. Notably, to reduce computational costs, we employ Low-Rank Adaptation \cite{lora} (LoRA) for fine-tuning Qwen2.5-VL. The fine-tuned model is subsequently utilized to extract high-quality captions from LR images, ensuring semantically coherent and visually grounded descriptions. As shown in Figure \ref{fig:caption}, fine-tuning enables the model to extract more precise captions from degraded LR images.

With a method capable of extracting high-quality captions from LR images, we further require the acquisition of foreground regions within the LR inputs. To this end, we employ LLMDet \cite{fu2025llmdet} for object detection on the LR images. Given the inherent robustness and computational efficiency of LLMDet, we refrain from fine-tuning it. As illustrated in Figure \ref{fig:ragsr} (a), the bounding boxes output by LLMDet and the original LR image are fed into the fine-tuned Qwen2.5-VL to generate region-specific captions and global captions. These prompts are then fed into the pre-trained T2I model to guide the super-resolution synthesis process.

\subsection{Regional Attention Control}

To enhance the model's capability to reconstruct fine-grained details within localized regions, we integrate regional attention control into both the training and inference phases of the T2I model, enabling alignment between detailed textual descriptions to their corresponding image regions. As illustrated in Figure \ref{fig:ragsr} (c), we design specialized regional attention masks for each input LR image to enforce a one-to-one correspondence between foreground image regions and their associated descriptions, while preventing cross-attention interactions between region-specific textual embeddings and semantically unrelated image regions.

Inspired by \cite{regional}, we divide the regional attention mask into four components: image-to-image self-attention mask $M_{i2i}$, text-to-text self-attention mask $M_{t2t}$, image-to-text cross-attention mask $M_{i2t}$ , and text-to-image cross-attention mask $M_{t2i}$. The representation is:

\begin{equation}
M_{\text{region}} = 
\begin{bmatrix} 
M_{t2t} & M_{t2i} \\ 
M_{i2t} & M_{i2i} 
\end{bmatrix}.
\end{equation}

For cross-attention from image to text, we apply regional masks to ensure that image tokens within each region attend solely to their corresponding text: 

\begin{equation}
M_{i2t} = \left[ \psi(R_h^1, R_c^1), \psi(R_h^2, R_c^2), \ldots, \psi(R_h^t, R_c^t) \right], 
\end{equation}
\begin{equation}
\psi(R_h^t, R_c^t) = R_h^t \otimes (R_c^t)^T.
\end{equation}

As illustrated in Figure \ref{fig:ragsr} (c), $R_h^t \in \{0, 1\}$ denotes the flattened regional mask associated with the $t$-th bounding box in the hidden states, while $R_c^t \in \{0, 1\}$ refers to $t$-th regional mask in the encoder hidden states. With the regional attention mask, our approach achieves alignment between textual descriptions and their corresponding visual regions. Specifically, in Figure \ref{fig:ragsr} (c), the visual features of the seagull region are restricted to attend solely to the regional captions: "A seagull in flight, wings spread wide against a cloudy sky".

Similarly, for the text-to-image cross-attention mask, we conveniently utilize 
$M_{t2i} = M_{i2t}^T$ to achieve correspondence between each description and specific image region.

Regarding the self-attention of image-to-image and text-to-text components, we execute correspondence through $R_h^t$, $R_c^t$ for each region and each description. Specifically, we define background region as $R_h^{\text{background}} = 1 - \text{LogicalOR}(R_h^1, \ldots, R_h^t)$. The components are calculated as:
\begin{equation}
M_{i2i} = \sum_{t}^{T} \left( R_h^t \otimes (R_h^t)^T \right) + R_h^{\text{background}} \otimes (R_h^{\text{background}})^T, 
\end{equation}
\begin{equation}
M_{t2t} = \sum_{t}^{T} \text{Region}(R_c^t) \otimes \text{Region}(R_c^t)^T.
\end{equation}

Thus, we obtain the latent for regional attention:
\begin{equation}
Z_{\text{region}} = \sigma(\text{ATTENTION}(Q_{\text{region}}, K_{\text{region}}, V_{\text{region}}, M_{\text{region}})), 
\end{equation}
$\sigma$ serves as a post-processing module for each transformer block. Notably, our approach avoids explicit fusion between the regional attention latent $Z_{\text{region}}$ and the global attention latent 
$Z_{\text{global}}$. We adopt a two-stage attention mechanism: first, extracting global attention features from the image, and then performing regional attention refinement on the extracted global features to enhance spatial-semantic alignment. This strategy enables the model to learn the relative contribution of global and regional features through end-to-end training, avoiding the need for manual weight tuning.

\section{Experiments}

\subsection{Experimental Setup}
\paragraph{Training.} 
We employ Qwen2.5-VL as our VLM for image caption generation. The training dataset is constructed by integrating four publicly available datasets: COCO-2017 Train \cite{coco2017}, Flickr30k-Entities \cite{flickr30k2015}, GQA \cite{gqa2019}, and LLaVA-Cap \cite{llava2024}, which ensures diverse and comprehensive visual-textual representation. To simulate real-world low-quality image inputs, we apply a degradation pipeline \cite{RealESRGAN2021} to the original images. For model adaptation, we implement Low-Rank Adaptation \cite{lora} (LoRA) with a rank of 8, while optimizing the training process using a learning rate of 1e-4. This setup ensures efficient adaptation while maintaining the pretrained model’s generalization capabilities. The final model is applied to generate accurate captions from low-quality images, bridging the performance gap between pristine and degraded visual input.

We adopt the pre-trained Stable Diffusion 2 model \cite{LDM2022} as the foundational framework, with ControlNet \cite{controlnet} integrated to regulate its generative process. The RAGSR model is trained on the LSDIR \cite{li2023lsdir} dataset and a subset of 10K facial images from FFHQ \cite{ffhq}. To generate LR-HR training pairs, we utilize the degradation pipeline from Real-ESRGAN \cite{RealESRGAN2021}. Training parameters include a batch size of 192, a learning rate of 5e-5, and a total of 10K training iterations.

We analyzed the number of bounding boxes detected by the LLMDet model for each image in the training set, finding that the average is approximately 4 and the median is 3. Based on this, we set the default number of bounding boxes per image to 5. For detected bounding boxes, we sort them by confidence scores and preliminarily filter out boxes with confidence scores less than 0.4 to obtain relatively accurate detection results. For the remaining detected boxes, if the number exceeds five, we take the top five boxes and their corresponding captions as inputs. If there are fewer than five boxes, we pad the missing bounding box coordinates with zeros and set the corresponding captions to empty strings.

\paragraph{Evaluation.} 

Following the methodology of previous works \cite{seesr2024,RealESRGAN2021}, we focus on the $\times$4 SISR task, although the proposed approach is applicable to various scaling factors. For comprehensive evaluation, we conduct experiments on three benchmark datasets: DIV2K-Val \cite{div2k}, RealSR \cite{realsr}, and DrealSR \cite{drealsr}. During inference, we implement 50-step DDIM sampling consistent with prior approaches and set the default number of bounding boxes per image to 5 to align with the training configuration. To ensure holistic performance assessment, we employ both reference-based and no-reference quality metrics. Reference-based fidelity measures include PSNR and SSIM \cite{ssim}. Reference-based perceptual quality measures include LPIPS3 \cite{lpips} and DISTS \cite{dists}. Distributional similarity between original and reconstructed images is evaluated via FID \cite{fid}. For no-reference image quality estimation, we apply NIQE \cite{niqe}, MANIQA \cite{maniqa}, MUSIQ \cite{musiq}, and CLIPIQA \cite{clipiqa} metrics.

\subsection{Main Results}

\begin{table}[]
\centering
\caption{
Quantitative comparison with state-of-the-art methods on the benchmarks. The best and second best results of each metric are highlighted in \textcolor{red}{\textbf{red}} and \textcolor{blue}{\underline{blue}}, respectively. All the metrics related to MMSR are derived from its paper and some metrics have not been evaluated since the code of MMSR has not been open-sourced.%
}
\small
\setlength{\tabcolsep}{2pt}
\renewcommand{\arraystretch}{1.15}
\resizebox{\linewidth}{!}
{%
\begin{tabular}{c|c|ccccccccc}
\toprule
Datasets & Methods & PSNR & SSIM & LPIPS $\downarrow$ & DISTS $\downarrow$ & FID $\downarrow$ & NIQE $\downarrow$ & MANIQA & MUSIQ & CLIPIQA \\
\midrule
\multirow{13}{*}{\begin{tabular}[c]{@{}c@{}}\textit{DIV2K-Val}\end{tabular}}
& BSRGAN \cite{BSRGAN2021} & 21.87 & 0.5539 & 0.4136 & 0.2737 & 64.28 & 4.7615 & 0.4834 & 59.11 & 0.5183 \\
& R-ESRGAN \cite{RealESRGAN2021} & \tbd{21.94} & \tbd{0.5736} & 0.3868 & 0.2601 & 53.46 & 4.9209 & 0.5251 & 58.64 & 0.5424 \\
& LDL \cite{LDL2022} & 21.52 & 0.5690 & 0.3995 & 0.2688 & 58.94 & 5.0249 & 0.5127 & 57.90 & 0.5313 \\
& DASR \cite{dasr2022} & 21.72 & 0.5536 & 0.4266 & 0.2688 & 67.22 & 4.8596 & 0.4346 & 54.22 & 0.5241 \\
& FeMASR \cite{chen2022femasr} & 20.85 & 0.5163 & 0.3973 & 0.2428 & 53.70 & \tbd{4.5726} & 0.4869 & 58.10 & 0.5597 \\
& LDM \cite{LDM2022} & 21.26 & 0.5239 & 0.4154 & 0.2500 & 41.93 & 6.4667 & 0.5237 & 56.52 & 0.5695 \\
& StableSR \cite{StableSR2024} & 20.84 & 0.4887 & 0.4055 & 0.2542 & 36.57 & 4.6551 & 0.5914 & 62.95 & 0.6486 \\
& ResShift \cite{yue2023resshift} & 21.75 & 0.5422 & 0.4284 & 0.2606 & 55.77 & 6.9731 & 0.5232 & 58.23 & 0.5948 \\
& PASD \cite{PASD2024} & 20.77 & 0.4958 & 0.4410 & 0.2538 & 40.77 & 4.8328 & 0.6049 & 66.85 & 0.6799 \\
& DiffBIR \cite{diffbir2024} & 20.94 & 0.4938 & 0.4270 & 0.2471 & 40.42 & 4.7211 & \tbd{0.6205} & 65.23 & 0.6664 \\
& SeeSR \cite{seesr2024} & 21.19 & 0.5386 & 0.3843 & 0.2257 & 31.93 & 4.9275 & 0.6198 & 68.33 & \tbd{0.6946} \\
& MMSR \cite{mmsr2025} & 21.74 & 0.5693 & \tbd{0.3707} & \trf{0.2071} & \trf{29.35} & \trf{4.2532} & -- & \trf{70.06} & \trf{0.7164} \\
& \textbf{RAGSR} & \trf{23.37} & \trf{0.5866} & \trf{0.3408} & \tbd{0.2107} & \tbd{31.44} & 4.6314 & \trf{0.6251} & \tbd{68.45} & 0.6874 \\
\midrule

\multirow{7}{*}{\begin{tabular}[c]{@{}c@{}}\textit{RealSR}\end{tabular}}
& R-ESRGAN \cite{RealESRGAN2021} & \trf{25.69} & \trf{0.7616} & \trf{0.2727} & \trf{0.2063} & 135.18 & 5.8295 & 0.5487 & 60.18 & 0.4449 \\
& StableSR \cite{StableSR2024} & 24.70 & 0.7085 & 0.3018 & \tbd{0.2135} & 128.51 & 5.9122 & 0.6221 & 65.78 & 0.6178 \\
& PASD \cite{PASD2024} & 24.29 & 0.6630 & 0.3435 & 0.2259 & 129.76 & \tbd{5.3628} & 0.6493 & 68.69 & 0.6590 \\
& DiffBIR \cite{diffbir2024} & 24.77 & 0.6572 & 0.3658 & 0.2310 & 128.99 & 5.5696 & 0.6253 & 64.85 & 0.6386 \\
& SeeSR \cite{seesr2024} & \tbd{25.18} & \tbd{0.7216} & 0.3009 & 0.2223 & \tbd{125.55} & 5.4081 & 0.6442 & 69.77 & 0.6612 \\
& MMSR \cite{mmsr2025} & 24.83 & 0.7003 & \tbd{0.2952} & -- & -- & -- & \tbd{0.6578} & \trf{71.33} & \tbd{0.6717} \\
& \textbf{RAGSR} & 24.88 & 0.7067 & 0.3082 & 0.2247 & \trf{119.79} & \trf{5.2013} & \trf{0.6611} & \tbd{70.00} & \trf{0.6894} \\
\midrule

\multirow{7}{*}{\begin{tabular}[c]{@{}c@{}}\textit{DrealSR}\end{tabular}}
& R-ESRGAN \cite{RealESRGAN2021} & \trf{28.64} & \trf{0.8053} & \trf{0.2847} & \trf{0.2089} & 147.62 & 6.6928 & 0.4907 & 54.18 & 0.4422 \\
& StableSR \cite{StableSR2024} & 28.13 & 0.7542 & 0.3315 & \tbd{0.2263} & 148.98 & 6.5354 & 0.5591 & 58.42 & 0.6206 \\ %
& PASD \cite{PASD2024} & 27.00 & 0.7084 & 0.3931 & 0.2515 & 159.24 & 5.8595 & 0.5850 & 64.81 & 0.6773 \\
& DiffBIR \cite{diffbir2024} & 26.76 & 0.6576 & 0.4599 & 0.2749 & 166.79 & \tbd{6.2935} & 0.5923 & 61.19 & 0.6346 \\
& SeeSR \cite{seesr2024} & 26.75 & 0.7405 & \tbd{0.3174} & 0.2315 & \tbd{147.39} & 6.3967 & 0.6052 & 65.09 & 0.6908 \\
& MMSR \cite{mmsr2025} & 27.28 & \tbd{0.7456} & 0.3249 & -- & -- & -- & \trf{0.6301} & \trf{68.93} & \trf{0.6999} \\
& \textbf{RAGSR} & \tbd{27.33} & 0.7433 & 0.3376 & 0.2385 & \trf{146.00} & \trf{6.2324} & \tbd{0.6189} & \tbd{65.92} & \tbd{0.6968} \\
\bottomrule
\end{tabular}
}
\label{tab:results}
\end{table}

\paragraph{Quantitative Comparisons.} 

We present a quantitative comparison of our method with other baselines on the DIV2K-Val, RealSR, and DrealSR datasets in Table \ref{tab:results}. The results show that:

Our method achieves state-of-the-art (SOTA) performance on the DIV2K-Val, RealSR, and DrealSR datasets, attaining the highest number of best and second best rankings across all metrics compared to existing baselines. Specifically, on the DIV2K-Val dataset, our approach demonstrates the best performance in four metrics: PSNR, SSIM, LPIPS, and MANIQA, with notable improvements of 6.5\% in PSNR and 8.1\% in LPIPS over the second best method. Additionally, it ranks the second best in three other metrics (DISTS, FID, and MUSIQ) and achieves the third best in two no-reference metrics (NIQE and CLIPIQA). On the RealSR and DrealSR datasets, while R-ESRGAN \cite{RealESRGAN2021} maintains superior fidelity and perceptual quality, it compromises image quality and visual realism. In contrast, our method preserves both fidelity and perceptual quality while achieving superior no-reference metric scores, effectively balancing these objectives. These results demonstrate that RAGSR excels in high-fidelity reconstruction, enhanced perceptual quality, and improved super-resolution image quality.

\begin{figure}[t]
    \centering
    \includegraphics[width=\linewidth]{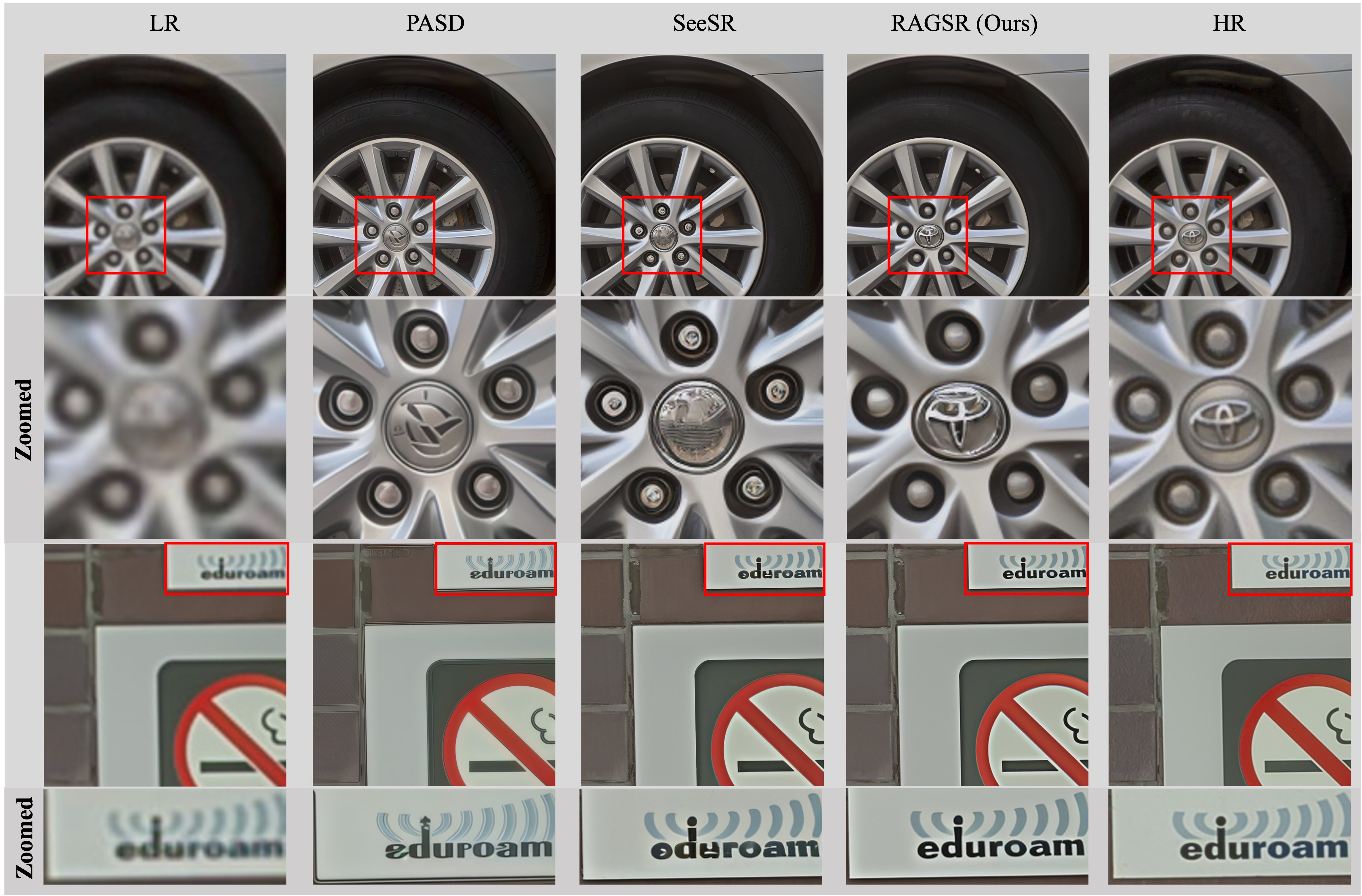}
    \caption{Qualitative comparisons of RAGSR and other SISR methods. RAGSR demonstrates excellence in global image quality and superiority in regional detail reconstruction compared to other methods.}
    \label{fig:qualitative}
\end{figure}

\paragraph{Qualitative Comparisons.}
Figure \ref{fig:qualitative} presents a comparison of the super-resolution results of our method with other baseline approaches. SeeSR \cite{seesr2024} utilizes tag-style prompts generated by a fine-tuned DAPE \cite{seesr2024} model. While it demonstrates reasonable performance in overall image reconstruction, it exhibits significant limitations in local detail restoration, such as failing to recover brand symbols on tires and the ``eduroam'' text on icons. PASD \cite{PASD2024}, which employs caption-style prompts, suffers from semantically incorrect text inputs, leading to degraded performance in both global and local detail reconstruction. In contrast, our approach leverages accurate global captions and region-specific captions extracted from the fine-tuned Qwen2.5-VL \cite{qwen252025} model. Under the guidance of a regional attention mechanism, these captions enable the generation of semantically coherent and detail-rich results.

\subsection{Analysis}
We conducted ablation experiments on our method using the RealSR dataset, encompassing three key aspects: (1) the impact of region control (presence/absence), (2) the effect of input prompt types (\eg, caption-style vs. tag-style), and (3) the influence of whether the Qwen2.5-VL model was fine-tuned on degraded images. The results are summarized in Table \ref{tab:results_prompt_region}. When regional attention control was applied and captions generated by the fine-tuned Qwen2.5-VL model were used as input, the model achieved the best performance. The detailed analysis is as follows:

\begin{table}[]
\centering
\caption{
Ablation study of RAGSR on the RealSR dataset. The asterisk (*) denotes captions generated using the fine-tuned Qwen2.5-VL model. The best and second best results of each metric are highlighted in \textcolor{red}{\textbf{red}} and \textcolor{blue}{\underline{blue}}, respectively. %
}
\small
\setlength{\tabcolsep}{2pt}
\renewcommand{\arraystretch}{1.15}
\resizebox{\linewidth}{!}
{%
\begin{tabular}{cc|ccccccccc}
\toprule
Prompt Type & Regional Control & PSNR & SSIM & LPIPS $\downarrow$ & DISTS $\downarrow$ & FID $\downarrow$ & NIQE $\downarrow$ & MANIQA & MUSIQ & CLIPIQA \\
\midrule
Caption* & \cmark & \trf{24.88} & \trf{0.7067} & \trf{0.3082} & \trf{0.2247} & \trf{119.79} & \tbd{5.2013} & 0.6611 & 70.00 & 0.6894 \\
Caption & \cmark & \tbd{24.86} & \tbd{0.7058} & 0.3112 & 0.2268 & 124.42 & 5.2717 & 0.6551 & 70.12 & 0.6838 \\
Tag & \cmark & 24.67 & 0.7022 & 0.3169 & 0.2328 & 128.85 & 5.4028 & \tbd{0.6649} & \tbd{70.94} & \tbd{0.6983} \\
Caption* & \xmark & 24.82 & 0.7041 & \tbd{0.3107} & \tbd{0.2261} & \tbd{120.56} & \trf{5.1757} & 0.6634 & 70.09 & 0.6897 \\
Caption & \xmark & 24.75 & 0.7035 & 0.3126 & 0.2281 & 124.57 & 5.2653 & 0.6578 & 70.34 & 0.6872 \\
Tag & \xmark & 24.64 & 0.6994 & 0.3185 & 0.2345 & 128.35 & 5.3900 & \trf{0.6668} & \trf{71.03} & \trf{0.7015} \\
\bottomrule
\end{tabular}
}
\label{tab:results_prompt_region}
\end{table}

\paragraph{Ablation of Regional Attention Control.}
By comparing rows in Table \ref{tab:results_prompt_region}, we observe that incorporating regional attention control improves the model's performance on four metrics—PSNR, SSIM, LPIPS, and DISTS, while slightly degrading on the other four metrics. However, the magnitude of improvement significantly outweighs the degradation. For instance, comparing the results in Row 1 and Row 4, the performance gain is 0.51\%, while the drop is only 0.25\%. This suggests that regional attention control achieves significant improvements in image fidelity and perceptual quality while incurring minimal degradation in image quality.

\begin{figure}[t]
    \centering
    \includegraphics[width=\linewidth]{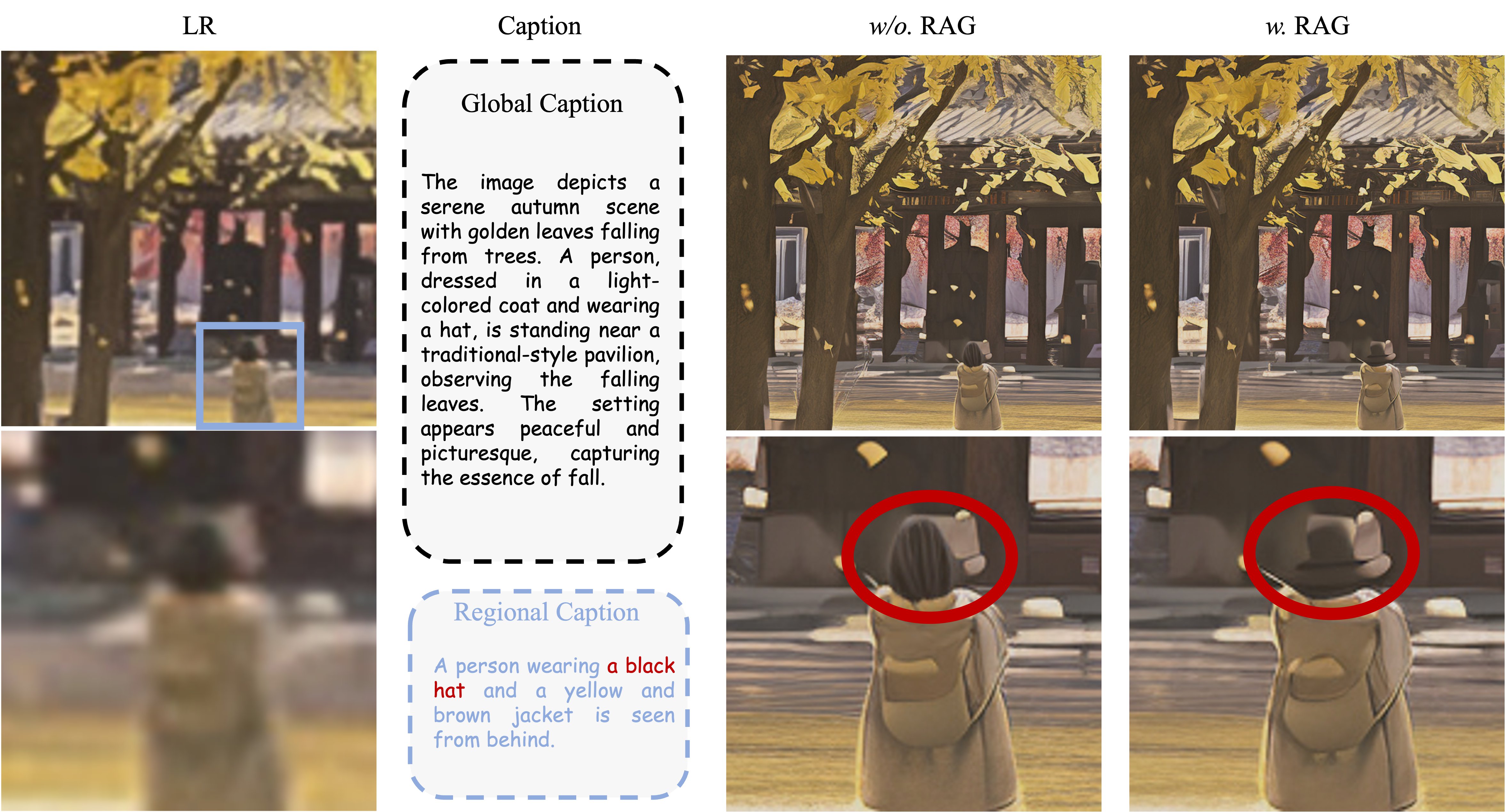}
    \caption{Ablation of regional attention control. Two identical RAGSR models are provided with the same global caption as a textual prompt, yet the model equipped with regional attention control successfully reconstructs the critical object ``hat'' during super-resolution, whereas the model without regional attention control fails to reconstruct ``hat'' despite its explicit mention in the global caption.}
    \label{fig:ablation}
\end{figure}

Figure \ref{fig:ablation} provides a more intuitive illustration of the role of regional attention control. As shown, when the same RAGSR model is applied with an identical global caption stating ``a person wearing a hat,'' the model without regional attention control fails to associate the textual cue with its corresponding region, resulting in the absence of the ``hat'' object in the super-resolution output. In contrast, the model incorporating regional attention control effectively maps the textual description to the designated spatial region, enabling accurate reconstruction of the ``hat'' in the final HR image. This demonstrates that regional attention control is critical for ensuring cross-modal alignment between text and image regions, particularly for preserving semantically important regional details.

\paragraph{Ablation of Prompt.}

By comparing the results in Table \ref{tab:results_prompt_region}, we observe the following:

The model achieves significant improvements in six objective quality metrics—PSNR, SSIM, LPIPS, DISTS, FID, and NIQE—when using captions instead of tags. However, there is a slight degradation in three image quality metrics: MANIQA, MUSIQ, and CLIPIQA. We hypothesize that using tags as input better enhances the model's generative capacity, as they provide concise and low-redundancy semantic cues, although the final SR outputs may deviate entirely from the intended content. However, captions offer richer scene descriptions, leading to more stable gains in image fidelity and perceptual quality, albeit at the cost of minor compromises in image quality.

Compared to captions generated by the original Qwen2.5-VL model, the performance of captions generated by the fine-tuned Qwen2.5-VL model is consistently better, regardless of whether regional attention control is incorporated. Fine-tuning Qwen2.5-VL systematically enhances the quality of SR across all metrics except MUSIQ, with the most notable gain (4.2\%) observed in FID. This indicates that fine-tuning significantly strengthens the model’s ability to extract accurate descriptions from LR images, aligning generated captions more closely with real-world semantic distributions. The overall performance validates the effectiveness of the fine-tuning strategy.

\section{Conclusion}
In this paper, we proposed RAGSR, a SISR method that utilizes regional captions as semantic priors to enhance the generative capability of pretrained T2I diffusion models. To minimize the influence of image degradation on semantic captions, we fine-tune Qwen2.5-VL on LR images to generate regional captions. By leveraging the region-text priors, we proposed the regional guided attention that ensures that each region-text pair is properly considered in the attention process while preventing unwanted interactions between unrelated region-text pairs. Our method achieves enhanced realism and accurate reconstruction, outperforming existing text-guided super-resolution methods both qualitatively and quantitatively.

There are some limitations of RAGSR. First of all, LLMDet may miss objects for heavily degraded images, resulting in imperfect results. Second, the fine-tuned Qwen2.5-VL may generate inaccurate text captions in cases of severe degradation. Third, while adding region-text information significantly enhances SR performance, it introduces computational overhead. 

\medskip

{
\small
\bibliographystyle{plain}
\bibliography{citation}
}




\newpage
\appendix
\section{Appendix}
\subsection{Additional Experiments}

We evaluate the performance of RAGSR on the real-world dataset RealLR200. Due to the absence of ground-truth high-resolution images in this dataset, we employ only no-reference image quality assessment metrics: NIQE, MANIQA, MUSIQ, and CLIPIQA. As shown in Table \ref{tab:reallr}, our method achieves the best scores across all metrics except for NIQE, with a notable 7.84\% improvement over the second best method on CLIPIQA. This indicates that the images generated by our method exhibit superior visual quality compared to existing approaches.

\begin{table}[h]
\centering
\caption{
Quantitative comparison with state-of-the-art methods on the real-world benchmark. The best and second best results of each metric are highlighted in \textcolor{red}{\textbf{red}} and \textcolor{blue}{\underline{blue}}, respectively.%
}
\small
\resizebox{0.65\linewidth}{!}
{%
\begin{tabular}{c|c|ccccccccc}
\toprule
Datasets & Methods & NIQE $\downarrow$ & MANIQA & MUSIQ & CLIPIQA \\
\midrule

\multirow{6}{*}{\begin{tabular}[c]{@{}c@{}}\textit{RealLR200}\end{tabular}}  
& R-ESRGAN  & 4.2048 & 0.5582 & 62.94 & 0.5389 \\
& StableSR  & 4.2516 & 0.5841 & 63.30 & 0.6068 \\
& PASD  & \tbd{4.1715} & 0.6066 & 68.20 & 0.6797 \\
& DiffBIR  & 4.9330 & 0.5902 & 62.06 & 0.6509 \\
& SeeSR  & \textcolor{red}{\textbf{4.1620}} & \tbd{0.6254} & \tbd{69.71} & \tbd{0.6813} \\
& \textbf{RAGSR}  & 4.2807 & \trf{0.6272} & \trf{71.30} & \trf{0.7347} \\
\bottomrule
\end{tabular}
}
\label{tab:reallr}
\end{table}

To investigate the impact of regional attention control injection at different inference time steps on the generated results, we conduct experiments on the RealSR dataset. Under a fixed total of 50 inference steps, we inject regional attention control at the first step, the first 10 steps, the first 25 steps, and all 50 steps, respectively. The model's performance is summarized in Table \ref{tab:ablation_steps}. The results show that increasing the number of injection steps enhances PSNR, SSIM, LPIPS, DISTS, and FID, while NIQE, MANIQA, MUSIQ, and CLIPIQA slightly deteriorate. As the number of injection steps for regional attention control increases, the generated images exhibit higher fidelity to the ground truth, achieving better reference-based perceptual quality. However, this improvement comes at the cost of reduced no-reference image quality. In this paper, the model defaults to 50 injection steps of regional attention control.  

\begin{table}[h]
\centering
\caption{
Ablation study on the injection steps of regional attention control in RAGSR. Model performance is evaluated on the RealSR dataset. The best and second best results of each metric are highlighted in \textcolor{red}{\textbf{red}} and \textcolor{blue}{\underline{blue}}, respectively. %
}
\small
\setlength{\tabcolsep}{2pt}
\renewcommand{\arraystretch}{1.15}
\resizebox{0.85\linewidth}{!}
{%
\begin{tabular}{c|ccccccccc}
\toprule
Injection Steps & PSNR & SSIM & LPIPS $\downarrow$ & DISTS $\downarrow$ & FID $\downarrow$ & NIQE $\downarrow$ & MANIQA & MUSIQ & CLIPIQA \\
\midrule
1 & 24.82 & 0.7043 & 0.3106 & 0.2260 & 120.48 & \trf{5.1836} & \trf{0.6633} & \trf{70.07} & \tbd{0.6893} \\
5 & 24.84 & 0.7050 & \tbd{0.3102} & 0.2258 & 120.54 & \tbd{5.1838} & \tbd{
0.6628} & \tbd{70.03} & 0.6886 \\
25 & \tbd{24.87} & \tbd{0.7064} & \trf{0.3082} & \tbd{0.2249} & \trf{119.73} & 5.1955 & 0.6614 & 69.98 & 0.6880 \\
50 & \trf{24.88} & \trf{0.7067} & \trf{0.3082} & \trf{0.2247} & \tbd{119.79} & 5.2013 & 0.6611 & 70.00 & \trf{0.6894} \\
\bottomrule
\end{tabular}
}
\label{tab:ablation_steps}
\end{table}

\subsection{Experimental Details}

\subsubsection{Training Details}

The training of both Qwen2.5-VL-7B and RAGSR models is conducted on 8 NVIDIA H20 GPUs. For Qwen2.5-VL, the prompt for generating image captions is set to "Please generate a very short description for the image." The initial learning rate, global batch size, and gradient accumulation steps are configured as 1e-4, 64, and 4, respectively, with the AdamW optimizer employed. For RAGSR, the initial learning rate, global batch size, and gradient accumulation steps are set to 5e-5, 96, and 2, respectively, using the Adam optimizer. To mitigate memory constraints during training, Low-Rank Adaptation (LoRA) is applied to both models, enabling efficient parameter updates with minimal additional memory overhead.

\subsubsection{Inference Speed}
We conduct a comprehensive analysis of the inference time composition of RAGSR. Specifically, the object detection speed of LLMDet averaged 3.59 images per second (img/s), while Qwen2.5-VL's caption extraction speed for images and corresponding bounding boxes averaged 0.22 img/s. The core model inference speed averaged 0.18 img/s. Compared to other text-driven super-resolution models, our method demonstrates competitive inference efficiency. For instance, the overall pipeline inference speed of RAGSR is approximately 20\% slower than PASD, but when bounding boxes and captions are pre-prepared, RAGSR's inference speed exceeds PASD by 30\%. Furthermore, RAGSR achieves superior performance in both image fidelity and visual quality compared to PASD. To ensure a fair comparison, all inference tasks are conducted on a single NVIDIA H20 GPU.

\subsection{Social Impact}

Our research introduces the integration of semantic priors from vision-language models (VLMs) and regional attention mechanisms with diffusion models, thereby driving innovative advancements in image super-resolution technology. This approach not only enables high-fidelity reconstruction of low-resolution images but also optimizes both perceptual quality and image quality metrics. This progress facilitates critical applications in medical imaging, satellite remote sensing, and archival restoration, where high-resolution details are indispensable. However, significant challenges remain: the method does not inherently address privacy risks associated with reconstructing sensitive visual information, nor does it resolve systemic biases or fairness concerns in the diffusion models and VLMs.

\subsection{More Samples}

\begin{figure}[h]
    \centering
    \includegraphics[width=\linewidth]{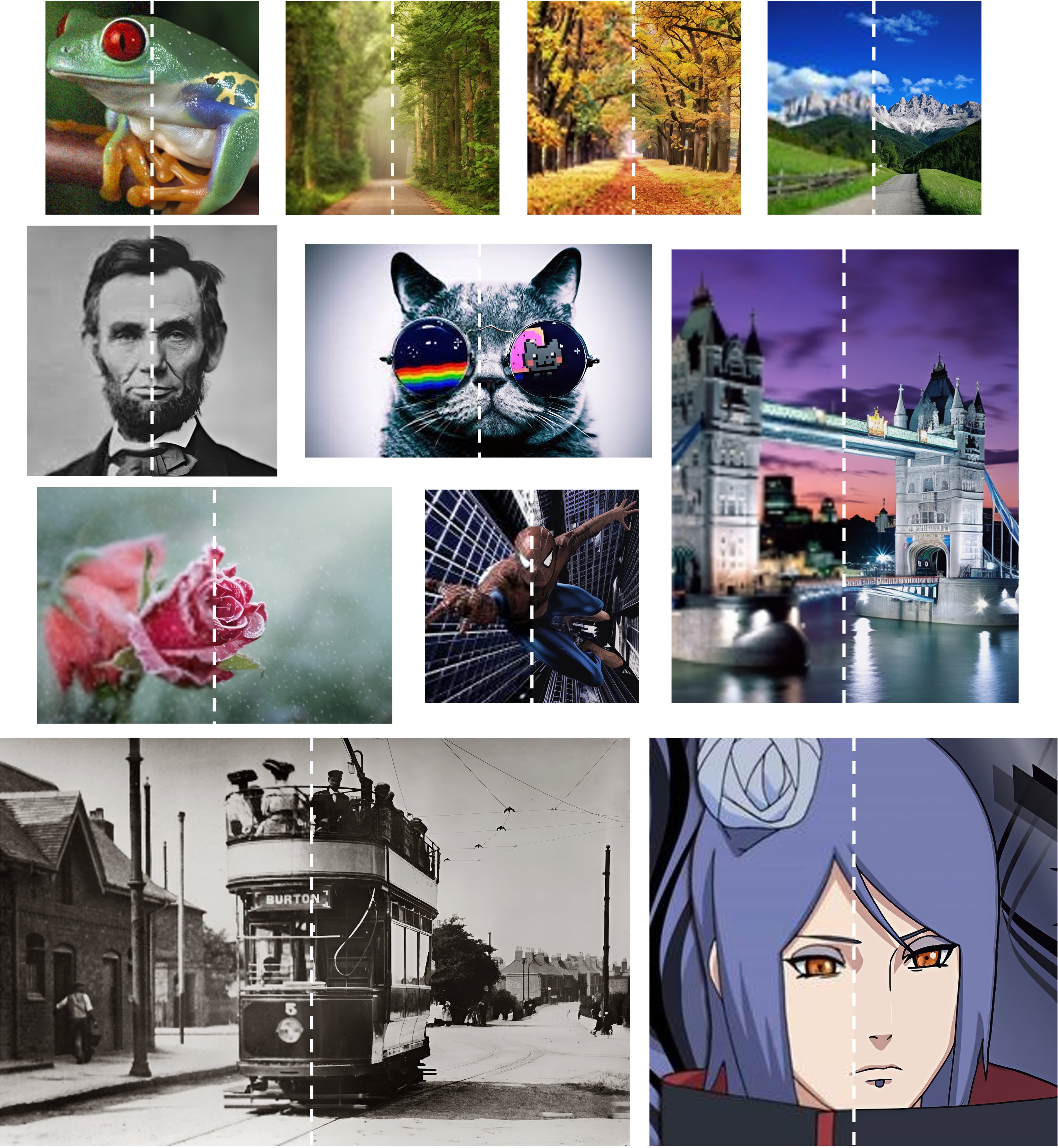}
    \caption{Super-resolution results of RAGSR on real-world images with varying resolutions.}
    \label{fig:samples}
\end{figure}

We present the 4$\times$ super-resolution performance of RAGSR on real-world images with various resolutions. As illustrated in Figure \ref{fig:samples}, the left half of each image (delimited by white dashed lines) represents the zoomed low-resolution input, while the right half displays the super-resolved results generated by our method. The images encompass monochrome and color modalities, with semantic content spanning human subjects, animals, flora, architecture, cartoons, and natural landscapes. The results clearly demonstrate that our approach achieves high-quality reconstructions while preserving fidelity to the original low-resolution content.

\end{document}